\documentclass[preprint,12pt]{elsarticle}

\usepackage{amssymb}
\usepackage{algorithm}
\usepackage{algorithmic}
\usepackage{xurl}
\usepackage{amsmath}
\usepackage{hyperref}
\usepackage{graphicx}
\usepackage{subcaption}
\usepackage{caption}
\usepackage{array}
\usepackage{booktabs}
\usepackage{tabularx}
\journal{Nuclear Physics B}

\begin{document}

\begin{frontmatter}



\title{A Vision-Based Framework Integrating Attention and Action Cues for Interpretable Cognitive Workload Assessment in Human--Robot Collaborative Assembly} 


\author[aff1,aff2]{Junyan Xiong}

\author[aff1]{Naiyi Feng}

\author[aff1]{Xingke Xia}

\author[aff1]{Qihang Fan}

\author[aff1]{Suchang Chen}

\author[aff1]{Daqiang Guo\corref{cor1}}

\cortext[cor1]{Corresponding author}

\affiliation[aff1]{
  organization={Smart Manufacturing Thrust, System Hub, The Hong Kong University of Science and Technology (Guangzhou)},
  city={Guangzhou},
  country={China}
}

\affiliation[aff2]{
  organization={College of Future Technology, The Hong Kong University of Science and Technology (Guangzhou)},
  city={Guangzhou},
  country={China}
}

\begin{abstract}
The introduction of human-robot collaboration (HRC) in industrial assembly operations is revolutionizing the manufacturing landscape. In this evolving environment, operators are required to seamlessly coordinate their manual tasks with real-time task information and robotic behaviors. These demands fluctuate during operation, yet conventional workload assessments depend on body-worn physiological sensors that complicate practical deployment. Here, we present a vision-based attention--action framework for continuous and interpretable workload-related assessment in HRC assembly. The framework combines RGB-D observations with robot states and calibrated task-related areas to construct a temporally confirmed representation of operator behavior. This representation identifies where task demand is concentrated and explains how it develops when attention and action diverge, the task context changes, or the operator hesitates. We evaluated the framework in a three-level collaborative gearbox assembly experiment with ten participants, using subjective ratings and synchronized physiological signals as independent references. Raw NASA-TLX ratings confirmed increasing perceived workload across conditions, with significant effects on overall workload and its mental and temporal dimensions. The vision-derived HRC-CWL output was significantly associated with ECG-derived features in seven of nine participants with complete correlation data. Synchronized interaction episodes further showed temporal correspondence between detected hesitation and physiological activity. Real-time deployment demonstrated that the framework can operate without requiring operators to wear additional sensors. These findings support HRC-CWL as an interpretable behavioral proxy for cognitive ergonomics analysis and adaptive robot assistance, rather than a direct psychophysiological measure of workload.
\end{abstract}

\begin{keyword}
Human--robot collaboration \sep Cognitive workload \sep Vision-based assessment \sep Attention--action modeling \sep Collaborative assembly
\end{keyword}



\end{frontmatter}



\section{Introduction}
\label{sec:introduction}
Industry 4.0 has been largely characterized by the adoption of advanced digital and automation technologies to improve production \cite{vogel2016guest}. However, this technology-oriented evolution has also raised concerns regarding the role of human workers in increasingly automated production systems \cite{XU2021530}. In response, Industry 5.0 has emerged as a complementary perspective that emphasizes industrial development around human needs and values \cite{LENG2022279,WANG2024102626}. Therefore human-centric manufacturing has gained increasing attention as a key principle of Industry 5.0, emphasizing the design of production systems and technologies that support and empower human operators in real factory settings \cite{XU2026259, LU2022612}.

Within the context of human centric manufacturing, human-robot collaboration (HRC) is deemed as an important manufacturing paradigm, as it enables the integration of human flexibility, perception, and decision-making with robotic precision, repeatability, and physical assistance \cite{LI2021547}. Robots are therefore no longer expected to simply replace human workers. Instead, they are increasingly required to support human operators in shared workspaces, while ensuring operator well-being \cite{YIN2023102515}. Among various manufacturing scenarios undergoing increasing automation and customization, assembly remains highly labor-intensive, as component variability and product customization continue to require human flexibility and expertise \cite{XIA2026103202}. Consequently, HRC systems have been increasingly adopted in assembly processes, where robots can support human workers in repetitive and physically demanding operations to enhance efficiency \cite{FATHI2024110254}. In a typical human--robot collaborative assembly (HRCA) scenario, humans and robots work side by side in close proximity within a shared workspace to perform shared tasks \cite{WANG2019701,LI2027103357}. 

Despite its potential benefits, HRCA may introduce significant ergonomic challenges, as frequent role switching, close-proximity interaction, and continuous coordination can increase operators' physical workload, cognitive strain, and even exposure to safety-related stressors \cite{GUALTIERI2020367}. Based on the principle of human-centricity, HRCA should prioritize operators' working conditions in production system design. This requires collaborative assembly systems to support human well-being while minimizing potential risks to operators' health, safety, and task performance \cite{10.3389/frobt.2022.813907}. Ergonomics in HRCA covers both physical and cognitive aspects of operators' well-being \cite{GUALTIERI2021101998}. Existing studies have long addressed physical ergonomics, mainly focusing on reducing biomechanical overload caused by repetitive motions, heavy handling, and awkward postures \cite{PARIGIPOLVERINI201725}. However, close collaboration also creates cognitive demands, especially when operators need to anticipate and respond to robot behavior \cite{8611391}. This makes cognitive ergonomics an important but still underexplored issue in collaborative assembly system design and optimization.

Cognitive ergonomics in collaborative assembly systems can be affected by several factors, including information overload, task complexity, robot motion, and stress \cite{10.1007/978-3-030-22216-1_21, GUALTIERI2022103807}. Excessive cognitive workload during collaborative tasks can negatively affect workers' performance and well-being. When work tasks and environments are poorly designed, they may impose high cognitive demands, leading to cognitive failures and potential harm to workers' mental health \cite{kalakoski2020effects}. To better understand operator states in the workspace, it is essential to evaluate the cognitive workload experienced by human operators during collaborative tasks. However, compared with physical ergonomics, the real-time assessment of cognitive workload in HRC assembly remains challenging. One reason is that cognitive workload is difficult to observe directly. Questionnaires such as NASA-TLX can provide reliable subjective evaluation, but they are usually collected after each task and cannot describe the continuous changes in operator state during task execution \cite{hart2006nasa}. Physiological signals such as ECG or EDA can provide additional objective evidence, but they require wearable devices and are easily affected by individual differences, sensor noise, and experimental setup \cite{10.3389/fphys.2019.00565, han2020objective}. 



To address these issues, this study proposes a vision-based attention--action framework for workload-related assessment in human--robot collaborative assembly. The framework models operator states within task-relevant regions of the shared workspace by jointly considering perceptual attention and action execution during collaboration. Based on visual observations and contextual information from the robot and task process, the proposed approach captures how operators allocate attention and perform actions under different collaborative conditions. These observations are further transformed into interpretable indicators, which characterize the underlying cognitive and interaction demands rather than relying solely on a single workload score.

The proposed proxy indicators are designed to capture workload-related patterns in collaborative assembly. For example, assembly execution demand reflects the attention invested in the assembly area, task search demand reflects part or tool searching behavior in the storage area, collaboration supervision demand reflects attention to the robot-related area, and context-aware attention switching demand reflects transitions among visual-action focus states. Overall hesitation demand reflects the operator's hesitation behavior during collaboration, indicating additional cognitive demands associated with interaction uncertainty and action preparation. These indicators are not intended to replace subjective workload measures. Instead, they provide continuous and interpretable evidence that can complement questionnaires and performance data. The main contributions of this study are summarized as follows:

\begin{enumerate}
    \item An HRC assembly-specific framework is established for real-time cognitive workload assessment by integrating operator states at both the attentional and behavioral levels within ongoing human--robot interactions.
    \item A lightweight vision-based workload assessment method is proposed for generating continuous workload-related metrics from an RGB-D stream without requiring participant-mounted sensors, while minimizing interference with natural human--robot collaboration.
    \item An interpretable workload decomposition mechanism is developed to transform evolving attention--action states into task- and interaction-related demand indicators, allowing workload variations to be traced to their behavioral sources and thereby supporting subsequent system-level optimization.
\end{enumerate}

The remainder of this paper is organized as follows. Section 2 reviews relevant studies on cognitive workload assessment in HRC assembly. Section 3 introduces the overall framework architecture and its processing workflow. Section 4 details the attention--action evaluation model and its workload-related indicators. Section 5 describes the experimental design and data collection. Section 6 presents the experimental evaluation and real-time implementation. Section 7 discusses the implications, practical relevance, and limitations of the findings. Section 8 concludes the paper and outlines future applications.

\section{Related Work}
\label{sec:relevantwirk}

Cognitive states in industrial tasks are commonly evaluated through objective physiological measurements, subjective questionnaires such as NASA-TLX, and performance-based metrics such as completion time, error rate, and task success \cite{attahir2025}. Objective methods are usually treated as a solid tool to analyze the human cognitive state during industrial settings. However, a single measurement type may be insufficient to reliably assess human cognitive states in industrial environments. Therefore, multimodal assessment has been increasingly adopted to integrate complementary information from different sources. For instance, Bussolan et al. \cite{10731373} proposed a multimodal stress detection framework for industrial human-robot collaborative assembly in which they used three physiological signals, including ECG, EMG, and EDA, together with facial action units and voice features. Their results showed that physiological signals provided strong discriminative information for stress recognition, while multimodal fusion further improved the robustness of operator state assessment in HRC scenarios. Manzardo et al. \cite{Manzardo_2026} established a more direct link between online assessment and robot adaptation in collaborative assembly. Their framework interprets gaze behavior as visual attention and pupil dilation as cognitive workload, then uses these two distinct operator-state proxies to regulate robot velocity and trajectory in real time.

However, despite their promising performance, such multimodal methods may face practical limitations in real industrial environments. The deployment of multiple wearable sensors and recording devices can increase operators' burden and may reduce the naturalness of assembly actions, particularly in industrial environments where signal quality is easily affected by movement and sensor instability. To overcome the deployment limitations of physiological and multimodal sensing methods, vision-based approaches have been introduced to support online cognitive load assessment in industrial tasks. Lagomarsino et al. \cite{9795908} proposed an online framework that estimates cognitive load from RGB-D camera data, using head pose estimation and skeleton tracking to identify workers' attention distribution and motion-related behavioral patterns. Instead of relying on wearable physiological devices, the framework derives interpretable indicators such as concentration loss, learning delay, instruction cost, self-touching, and hyperactivity. These indicators are further combined to estimate mental effort and stress level during assembly tasks, and the results were validated against physiological measures and subjective questionnaires. This work demonstrates the potential of non-intrusive visual sensing for practical cognitive load assessment in manufacturing environments.

Although multimodal physiological approaches provide valuable evidence for cognitive workload assessment, their reliance on complex sensor setups and time-consuming synchronized data collection limits their implementation in real assembly settings. Vision-based methods can therefore provide a more readily deployable alternative by inferring cognitive workload from observable human behavior. Their direct application to HRC, however, remains limited because existing frameworks are largely developed for conventional assembly and often define workload indicators based on behaviors, such as self-touching, that are not representative of collaborative task execution. This mismatch becomes particularly evident when robot induces attention reallocation and introduces extra cognitive demands. In addition, representing workload as a general score obscures its underlying sources and provides limited guidance for determining which aspects of the task or robot behavior should be adapted.

Together, these limitations call for a workload assessment approach that is tailored to HRC while remaining timely, unobtrusive, and diagnostically informative. The approach should capture cognitive demands associated with robot-induced shifts in attention and continuous coordination, which conventional manual-assembly indicators may overlook. Because these demands evolve throughout task execution and interaction, they must be assessed early enough to enable adaptation during operation. The assessment should also impose minimal sensing and procedural burdens to preserve natural operator behavior in shared workspaces. Moreover, interpretable proxies should distinguish workload arising from task execution from that induced by human--robot interaction, thereby providing actionable guidance for selecting appropriate adaptations. An approach that does not meet these requirements is therefore unlikely to provide the timely and actionable evidence needed for practical adaptation in HRC.

Table \ref{tab:objective_measurements} positions these and other representative approaches against the capabilities emphasized in this study: applicability to HRC, real-time assessment, lightweight deployment, and interpretable proxy outputs that can inform subsequent adaptation. Across existing work, real-time and lightweight methods generally lack either HRC applicability or interpretable proxies, while HRC-oriented approaches often depend on non-lightweight sensing or operate offline. Consequently, no surveyed approach jointly provides all four capabilities needed to translate continuous assessment into practical robot adaptation. To bridge this gap, this study proposes a vision-based attention--action framework for interpretable cognitive workload assessment in human--robot collaborative assembly.


\begin{table}[H]
\centering
\caption{Comparison of representative cognitive workload assessment approaches and the proposed framework. A solid circle ($\bullet$) denotes an explicitly demonstrated capability, whereas a hollow circle ($\circ$) denotes that the capability was not reported.}
\label{tab:objective_measurements}
\scriptsize
\setlength{\tabcolsep}{4pt}
\renewcommand{\arraystretch}{1.2}
\begin{tabularx}{\textwidth}{
    @{}
    >{\raggedright\arraybackslash}X
    >{\centering\arraybackslash}p{0.19\textwidth}
    >{\centering\arraybackslash}p{0.12\textwidth}
    >{\centering\arraybackslash}p{0.14\textwidth}
    >{\centering\arraybackslash}p{0.20\textwidth}
    @{}}
\toprule
\parbox[c][3.2\baselineskip][c]{\linewidth}{\raggedright\textbf{Approach}}
& \parbox[c][3.2\baselineskip][c]{\linewidth}{\centering\textbf{HRC}\\\textbf{applicability}}
& \parbox[c][3.2\baselineskip][c]{\linewidth}{\centering\textbf{Real-time}\\\textbf{assessment}}
& \parbox[c][3.2\baselineskip][c]{\linewidth}{\centering\textbf{Lightweight}\\\textbf{deployment}}
& \parbox[c][3.2\baselineskip][c]{\linewidth}{\centering\textbf{Interpretable}\\\textbf{proxies for}\\\textbf{adaptation}}
\tabularnewline
\midrule
Lagomarsino et al. \cite{9795908} & $\circ$ & $\bullet$ & $\bullet$ & $\circ$ \tabularnewline
Bussolan et al. \cite{10731373} & $\bullet$ & $\bullet$ & $\circ$ & $\circ$ \tabularnewline
Memar and Esfahani \cite{10.1145/3368854} & $\circ$ & $\circ$ & $\circ$ & $\circ$ \tabularnewline
Zhao et al. \cite{Zhao_2024} & $\bullet$ & $\bullet$ & $\bullet$ & $\circ$ \tabularnewline
Pluchino et al. \cite{10.3389/frobt.2023.1275572} & $\bullet$ & $\circ$ & $\circ$ & $\circ$ \tabularnewline
Manzardo et al. \cite{Manzardo_2026} & $\bullet$ & $\bullet$ & $\circ$ & $\bullet$ \tabularnewline
Zakeri et al. \cite{s23218926} & $\bullet$ & $\circ$ & $\circ$ & $\circ$ \tabularnewline
Xue et al. \cite{Xue_2026} & $\bullet$ & $\circ$ & $\circ$ & $\bullet$ \tabularnewline
Kchour et al. \cite{Kchour_2026} & $\circ$ & $\bullet$ & $\circ$ & $\circ$ \tabularnewline
\midrule
\textbf{Ours} & $\bullet$ & $\bullet$ & $\bullet$ & $\bullet$ \tabularnewline
\bottomrule
\end{tabularx}
\vspace{2pt}
\parbox{\textwidth}{\scriptsize\emph{Note:} HRC applicability indicates that the approach was developed or evaluated in a human--robot collaborative task. Lightweight deployment denotes assessment without participant-mounted physiological or eye-tracking sensors or dedicated immersive hardware. Interpretable proxies refer to semantically distinct outputs that identify a potential target for downstream adaptation.}
\end{table}

\section{Overall Framework of Cognitive Assessment in HRC}
\label{sec:framework}

The proposed framework targets workload-related behavioral assessment in human--robot collaborative assembly, where a UR5e robot supports component delivery and handover. Rather than estimating a single cognitive workload score, it jointly characterizes attention allocation and action behavior.

\begin{figure*}[t]
    \centering
    \includegraphics[width=\textwidth]{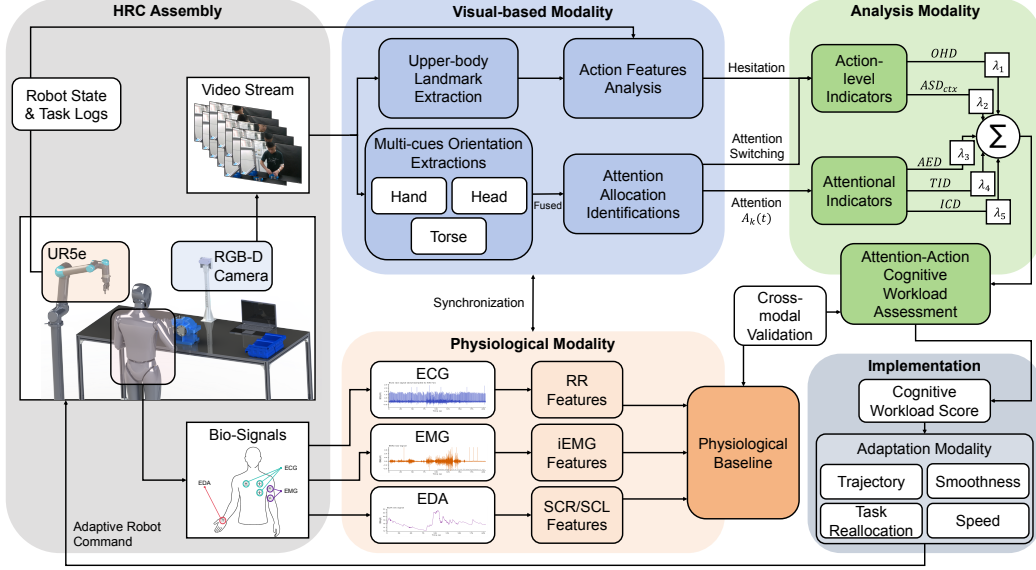}
    \caption{Overall framework for vision-based attention--action workload
    assessment in human--robot collaborative assembly. Visual observations are
    transformed into attentional and interaction-related proxy indicators,
    evaluated against physiological responses, and used to support adaptive
    robot behavior.}
    \label{fig:framework}
\end{figure*}

As illustrated in Figure~\ref{fig:framework}, the framework comprises four components: an HRC assembly task layer that defines the collaborative context; a vision-based modality that extracts attention and action characteristics from an RGB-D stream; an analysis modality that derives interpretable indicators of cognitive and interaction demands; and a validation and implementation layer that evaluates these indicators using physiological measurements and explores their use in adaptive robot assistance.


\subsection{Human--Robot Collaborative Assembly and Data Inputs}
\label{subsec:framework_inputs}

The collaborative assembly task layer provides the task context required for interpreting human behaviors during HRC. Unlike conventional assembly scenarios where human actions can be analyzed independently, HRC requires continuous coordination between task execution and robot assistance. Therefore, the proposed framework incorporates task-related information, including assembly phases, robot interaction events, and execution states, to establish the context in which visual behaviors occur.

The collaborative task is designed with three complexity levels to induce different cognitive and interaction demands. The increasing complexity is introduced through variations in component handling, assembly sequence constraints, robot-assisted operations, and monitoring requirements. These task conditions provide diverse interaction scenarios for evaluating whether the proposed attention-action indicators can capture workload-related changes during collaborative assembly.

During task execution, the robot state and task logs are synchronised with the visual data stream. Robot motion states provide information about interaction timing and robot-related events, while task logs describe the current execution stage. This contextual information is subsequently used to interpret human attention and action patterns with respect to the ongoing collaboration process.

\subsection{Visual-Based Modality}
\label{subsec:visual_modality}

The visual-based modality uses synchronized RGB-D observations to describe
operator behavior in the collaborative workspace.
Its processing distinguishes perceptual orientation from physical engagement,
while allowing both forms of evidence to support a common area-state
representation.

Upper-body and hand landmarks are tracked to determine how the operator
reorients and whether manual activity continues during task execution. The
resulting trajectories describe physical engagement and reveal sustained
low-motion behavior within robot-related collaboration periods.

Perceptual orientation is estimated primarily from head direction. Upper-body
orientation supports this estimate during large turns or unreliable head
observations, whereas hand position indicates where physical action is taking
place. The model therefore assigns different roles to these cues instead of
treating them as equivalent estimates of attention.

Each cue is evaluated against calibrated workspace areas whose functions are
defined by the assembly context. The available evidence is combined according
to its reliability without forcing weak or ambiguous observations into an area.
Temporal confirmation then suppresses frame-level fluctuations and yields the
stable area state used in the subsequent demand analysis.

The model retains perceptual orientation and action engagement separately after
forming the area state. It can therefore identify situations in which the
operator monitors instructions or the robot while continuing an assembly action
elsewhere. Together with hand-motion information and synchronized interaction
context, these states form the input to the analysis modality.

\subsection{Analysis Modality}
\label{subsec:analysis_modality}

The analysis modality converts visual observations into attentional and
interaction-related proxy indicators. These indicators characterize observable
workload-related behavior rather than directly measuring the operator's internal
cognitive state.

\subsubsection{Attentional Indicators}
\label{subsubsec:attentional_indicators}

The attentional indicators quantify accumulated attention across task-relevant
regions. Assembly execution demand (AED) represents attention associated with
assembly operations, tool identification demand (TID) with storage or component
selection, and instruction checking demand (ICD) with task instructions.
Together, they identify the predominant source of visually observable task
demand.

\subsubsection{Interaction-Related Indicators}
\label{subsubsec:interaction_indicators}

Context-aware attention switching demand, $ASD_{\mathrm{ctx}}$, describes
stable attention reallocations during human--robot collaboration. A minimum
dwell requirement filters brief focus fluctuations, after which retained
transitions are weighted by the functional relationship between their source
and destination regions. The indicator therefore reflects both transition
frequency and contextual cost rather than simply counting AOI transitions.

Overall hesitation demand (OHD) describes sustained low-motion hand behavior
during robot-related collaboration. Smoothed trajectories are evaluated using
adaptive velocity and duration thresholds, and detected intervals are
accumulated relative to the collaboration episode duration. Restricting the
calculation to active collaboration distinguishes hesitation-like behavior from
general task inactivity.

The five indicators are integrated into an overall workload-related assessment
but retained individually to preserve interpretability. Their mathematical
definitions and parameter settings are provided in
Section~\ref{sec:visual_model}.

\begin{table}[H]
\centering
\caption{Vision-derived workload-related proxy indicators.}
\label{tab:proxy_indicators}
\small
\setlength{\tabcolsep}{6pt}
\renewcommand{\arraystretch}{1.15}
\begin{tabularx}{\textwidth}{
    @{}
    >{\raggedright\arraybackslash}p{0.12\textwidth}
    >{\raggedright\arraybackslash}X
    >{\raggedright\arraybackslash}p{0.29\textwidth}
    @{}}
\toprule
\textbf{Indicator}
& \textbf{Visual basis and interpretation}
& \textbf{Typical conditions}
\tabularnewline
\midrule

\textbf{AED}
& Attention assigned to the assembly area, representing visually observable demand during task execution.
& Prolonged or spatially complex assembly operations.
\tabularnewline
\addlinespace[3pt]

\textbf{TID}
& Attention assigned to the storage area, representing demand associated with locating and selecting task components.
& Several similar or ambiguously located components.
\tabularnewline
\addlinespace[3pt]

\textbf{ICD}
& Attention assigned to the instruction area, representing demand associated with consulting task guidance.
& Frequent or prolonged instruction checking.
\tabularnewline
\addlinespace[3pt]

$\boldsymbol{ASD}_{\mathrm{ctx}}$
& Confirmed transitions between attention states, representing contextual demand during attention reallocation.
& Frequent switching across task and robot-related areas.
\tabularnewline
\addlinespace[3pt]

\textbf{OHD}
& Sustained low-motion hand behavior detected within robot-collaboration periods, representing interaction-related hesitation.
& Uncertain handover timing or greater interaction difficulty.
\tabularnewline

\bottomrule
\end{tabularx}
\end{table}

\subsection{Physiological Modality and Validation}
\label{subsec:physiological_validation}

The physiological modality provides an independent reference for examining the
workload-related relevance of the visually derived indicators. ECG, EMG, and
EDA signals are synchronously acquired during task execution. ECG is processed
to obtain RR-interval and heart-rate-variability features \cite{tao2019systematic}, EMG is transformed into integrated muscle-activity features, and EDA is decomposed into tonic and
phasic components represented by skin conductance level and skin conductance
responses.

These physiological features are not incorporated into the vision-based
assessment during model inference. Instead, they are used to examine whether
the visually derived indicators exhibit changes consistent with physiological
responses across different collaborative assembly conditions \cite{8115233}. The monotonic association between a visual indicator and a physiological feature is expressed as

\begin{equation}
\rho_s
=
\operatorname{corr}
\left(
\operatorname{rank}(X),
\operatorname{rank}(Y)
\right),
\label{eq:spearman_validation}
\end{equation}

where $X$ denotes a visually derived indicator and $Y$ denotes the corresponding
physiological feature. The detailed statistical procedure and significance
analysis are presented in Section~\ref{sec:experimental_results}.

This validation strategy examines whether the proposed visual proxies covary
with physiological responses without treating any single physiological feature
as an absolute ground-truth measurement of cognitive workload. The
physiological modality therefore serves as an external validation reference
rather than an additional input modality for online visual inference.

\subsection{Implementation and Robot Adaptation}
\label{subsec:implementation}

The framework produces both an overall workload-related output and its
underlying attentional and interaction-related components. This decomposition
enables the system to identify not only whether workload-related behavior
increases, but also whether the observed change is predominantly associated
with assembly execution, instruction consultation, component identification,
attention switching, or interaction hesitation.

The assessment can subsequently support adaptive robot behavior. Depending on
the dominant indicator, the robot may modify its motion speed, trajectory
smoothness, handover timing, or task allocation. For example, a sustained rise
in interaction-related hesitation may motivate a slower or smoother handover,
whereas frequent attention switching may indicate that the interaction sequence
or information presentation should be simplified.

The adaptation module is treated as a downstream implementation of the
assessment framework rather than as part of the visual indicator definition.
This separation allows the assessment model to remain interpretable while
supporting different robot adaptation strategies according to the requirements
of the collaborative task.

\section{Attention-Action Evaluation Model for HRC Assembly}
\label{sec:visual_model}



Human--robot collaborative assembly requires operators to coordinate visual attention, manual action, and robot-related events within the same workspace. In HRC, workload arises not only from task execution but also from continuous coordination with the robot. The proposed model therefore treats cognitive workload-related demand as an observable coordination problem rather than as a directly measured internal state. It first represents how the operator is oriented toward and physically engaged with task-related areas, and then derives demand descriptors from the resulting area-state sequence. This design links visual cue extraction, area-state estimation, and HRC-specific demand indicators within a single interpretable pipeline.

\subsection{Model Overview}

The model operationalizes the proposed assessment pipeline using the RGB-D observation sequence $V$, robot states $R$, calibrated task-related areas, and camera parameters $K$. These inputs are transformed into an interpretable attention--action cognitive workload representation, denoted as $\mathrm{HRC\text{-}CWL}(t)$, which describes workload-related changes during collaborative assembly.

For each frame, head, upper-body, and hand cues are mapped to functional regions of the collaborative workspace. Reliability-aware fusion combines the available visual evidence, while temporal confirmation suppresses transient fluctuations and produces a stable area state. The resulting state sequence provides a common basis for characterizing where task demand is concentrated and how attention--behavior coordination evolves during human--robot interaction.

This process begins by establishing a shared spatial reference. The collaborative workspace is partitioned into functionally distinct task-related areas:

\begin{equation}
\mathcal{A}
=
\left\{
A_1,A_2,\ldots,A_N
\right\},
\label{eq:area_set}
\end{equation}

where $A_i$ denotes the $i$th area associated with a specific task context. In the implemented collaborative assembly scenario, $A_{\mathrm{ass}}$, $A_{\mathrm{id}}$, $A_{\mathrm{ins}}$, and $A_{\mathrm{col}}$ represent the assembly, tool-identification, instruction, and robot-collaboration areas, respectively. These calibrated areas connect the estimated perceptual orientation and physical engagement of the operator with the subsequent workload-related metrics.

\subsubsection{Cue-specific Area Evidence Mapping}

At time $t$, the model extracts head $q_h(t)$, upper-body $q_b(t)$, and hand $q_m(t)$ cues as complementary visual evidence. In the present implementation, these cues are represented by the geometric states $q_h(t)=\{\mathbf{o}_h(t),\mathbf{x}_h(t)\}$, $q_b(t)=\{\mathbf{o}_b(t),\mathbf{x}_b(t)\}$, and $q_m(t)=\{\mathbf{x}_m(t)\}$. Here, $\mathbf{o}_h(t)$ and $\mathbf{o}_b(t)$ are unit orientation vectors, while $\mathbf{x}_h(t)$, $\mathbf{x}_b(t)$, and $\mathbf{x}_m(t)$ are the estimated head, upper-body, and hand positions. The head cue provides the primary proxy for visual attention. The upper-body and hand cues provide supporting behavioral evidence when the operator is turning between areas or physically engaged with an object.

The three cues have different units and interpretations, so each cue is first expressed as a deviation from every task-related area. Let $\mathbf{r}_i$ denote the representative center of area $A_i$. For cue $c\in\{h,b,m\}$, the deviation from area $A_i$ is defined as

\begin{equation}
\delta_i^c(t)
=
\begin{cases}
\angle
\left(
\mathbf{o}_h(t),
\mathbf{r}_i-\mathbf{x}_h(t)
\right),
&
c=h,
\\[4pt]
\angle
\left(
\mathbf{o}_b(t),
\mathbf{r}_i-\mathbf{x}_b(t)
\right),
&
c=b,
\\[4pt]
\operatorname{dist}
\left(
\mathbf{x}_m(t),A_i
\right),
&
c=m.
\end{cases}
\label{eq:cue_area_deviation}
\end{equation}

For head and body cues, $\delta_i^c(t)$ measures angular misalignment with area $A_i$. For the hand cue, it measures spatial distance to that area. This step does not determine the final area state. It only expresses how strongly each cue supports each candidate area.

Because the angular and spatial deviations have different units, they are normalized by their physical ranges:

\begin{equation}
\widetilde{\delta}_i^c(t)
=
\begin{cases}
\dfrac{\delta_i^c(t)}{\pi},
& c\in\{h,b\},
\\[6pt]
\dfrac{\delta_i^c(t)}{D_{\mathrm{ws}}},
& c=m,
\end{cases}
\label{eq:normalized_cue_area_deviation}
\end{equation}

where the angular deviations are expressed in radians and $D_{\mathrm{ws}}>0$ is the diagonal length of the calibrated workspace. The normalized deviation is then converted into a cue-specific area evidence score:

\begin{equation}
S_i^c(t)
=
\max
\left\{
0,
1-\widetilde{\delta}_i^c(t)
\right\},
\qquad
c\in\{h,b,m\}.
\label{eq:cue_area_evidence}
\end{equation}

The linear mapping assigns stronger evidence to smaller deviations, while the maximum operator prevents distances beyond the calibrated range from producing negative scores. Thus, $S_i^c(t)\in[0,1]$ can be compared across cues without introducing a tunable scale parameter.

\subsubsection{Multi-cue Fusion and Orientation Area Identification}

The cue-specific scores are then combined within each task-related area. The fusion uses reliability-aware adaptive weights:

\begin{equation}
S_i(t)
=
\sum_{c\in\{h,b,m\}}
\omega_c(t) S_i^c(t),
\qquad
S_i(t)\in[0,1].
\label{eq:multi_cue_fusion}
\end{equation}

where $\omega_c(t)$ denotes the reliability weight of cue $c$ at time $t$, with the cue weights normalized to sum to one. The weighting strategy follows the role of each cue. Under normal observations, head orientation receives the largest weight because it is the closest available proxy for visual attention. Upper-body orientation and hand motion provide supporting evidence. If the head cue is missing or has low confidence, its contribution is reduced and the remaining cues are strengthened. During a large upper-body reorientation, the body cue is allowed to dominate because it better represents an ongoing transition between task-related areas. The resulting $S_i(t)$ is an orientation score rather than a probability. It indicates the strength of fused support for area $A_i$ at the current frame.

The fused scores are converted into a stable area state through a two-stage update. First, the dominant current-frame area is identified as $i_S^{*}(t)=\arg\max_{i\in\{1,\ldots,N\}}S_i(t)$. The raw state $F_{\mathrm{raw}}(t)$ is assigned to $A_{i_S^{*}(t)}$ only when $S_{i_S^{*}(t)}(t)\geq\tau$; otherwise, it is treated as unassigned. The raw sequence is then summarized over a short temporal window to reduce frame-level fluctuations:

\begin{equation}
\eta_i(t)
=
\frac{1}{M}
\sum_{k=0}^{M-1}
\mathbb{I}
\left[
F_{\mathrm{raw}}(t-k)=A_i
\right],
\qquad
\eta_i(t)\in[0,1].
\label{eq:temporal_membership}
\end{equation}

The confirmed area state is updated only when the strongest temporal membership exceeds a confirmation threshold:

\begin{equation}
F(t)
=
\begin{cases}
A_{i_\eta^{*}(t)},
&
\eta_{i_\eta^{*}(t)}(t)\geq\gamma,
\\[4pt]
F(t-1),
&
\eta_{i_\eta^{*}(t)}(t)<\gamma.
\end{cases}
\label{eq:temporal_confirmation}
\end{equation}

where $i_\eta^{*}(t)=\arg\max_{i\in\{1,\ldots,N\}}\eta_i(t)$. The frame-level threshold $\tau$ prevents weak or ambiguous evidence from being forced into a task-related area. The temporal confirmation threshold $\gamma$ specifies the minimum fraction of observations within the $M$-frame window that must support the same area before the confirmed state is updated. A strict-majority criterion is adopted, requiring more than half of the window observations to support the candidate area. The resulting $F(t)$ is the stable area state used by the demand indicators below.

\subsection{Task-related Demand and Attention--Behavior Inconsistency}
\label{subsec:task_demand_abi}

\subsubsection{Area-based Task Demand}

The confirmed area state provides the basis for the first group of demand descriptors. Task-related demand is described by how valid observation time is distributed across the calibrated task areas \cite{EVANS2020104119}. Before accumulating these durations, we define a binary validity mask $v_F(u)\in\{0,1\}$ for the fused-state evaluation. Specifically, $v_F(u)=1$ when the fused state can be evaluated from the available observations at time $u$, whereas $v_F(u)=0$ when the observation is missing or invalid. The unassigned outcome $F(u)=\varnothing$ remains a valid evaluated state and therefore does not, by itself, set $v_F(u)$ to zero. For an arbitrary area $A_i$, the accumulated occupation time over the interval $[0,t]$ is then defined as

\begin{equation}
T_i(t)
=
\int_0^t
\mathbb{I}
\left[
F(u)=A_i
\right]
v_F(u)
\,du,
\label{eq:area_occupation_time}
\end{equation}

The corresponding accumulated valid observation time is

\begin{equation}
T_v(t)
=
\int_0^t
v_F(u)
\,du.
\label{eq:valid_observation_time}
\end{equation}

The accumulated valid observation time provides the denominator for the following duration-normalized demand measure. Accordingly, $D_i(t)$ is initialized to zero and updated only when $T_v(t)>0$, while values obtained before this condition are not interpreted as valid assessments. The same initialization convention is applied to the subsequent duration-normalized indicators. Under this convention, the demand associated with area $A_i$ is formulated as

\begin{equation}
D_i(t)
=
\frac{
T_i(t)
}{
T_v(t)
},
\label{eq:general_area_demand}
\end{equation}

which represents the proportion of valid observation time during which the operator's fused attention-orientation state is assigned to area $A_i$.

Applying Eq.~\eqref{eq:general_area_demand} to each calibrated area converts the generic measure $D_i(t)$ into a task-context-specific descriptor. For the assembly, identification, and instruction areas, the resulting descriptors are reported as assembly execution demand (AED), tool identification demand (TID), and instruction checking demand (ICD), respectively. Applying the same measure to the robot-collaboration area characterizes the proportion of valid observation time allocated to robot-related coordination. These descriptors therefore share the same computational basis and differ only in the functional context represented by each area. They describe the temporal allocation of observable attention--action states rather than directly measuring cognitive workload.

\subsubsection{Attention--Behavior Inconsistency}

Area occupation captures where the operator is mainly oriented. However, it does not distinguish coordinated behavior from cases in which perception and action are directed to different task areas. The perceptual and action components are therefore retained separately to characterize attention--behavior inconsistency \cite{VANKAMPEN201942}.

The perceptual orientation state is defined as

\begin{equation}
F^{\mathrm{per}}(t)
=
\Psi_{\mathrm{per}}
\left(
q_h(t),
q_b(t),
\mathcal{A}
\right),
\label{eq:perceptual_state}
\end{equation}

where $q_h(t)$ and $q_b(t)$ denote head- and upper-body-related cues.

The action engagement state is defined as

\begin{equation}
F^{\mathrm{act}}(t)
=
\Psi_{\mathrm{act}}
\left(
q_m(t),
\mathcal{A}
\right),
\label{eq:action_state}
\end{equation}

where $q_m(t)$ denotes the hand-related cue.

The instantaneous attention--behavior inconsistency is

\begin{equation}
I_{\mathrm{AB}}(t)
=
\mathbb{I}
\left[
F^{\mathrm{per}}(t)
\neq
F^{\mathrm{act}}(t)
\right]
v_{\mathrm{AB}}(t),
\label{eq:ab_inconsistency_state}
\end{equation}

where $v_{\mathrm{AB}}(t)\in\{0,1\}$ indicates that both sub-states are valid.

The accumulated attention--behavior inconsistency index is defined as

\begin{equation}
\mathrm{ABI}(t)
=
\frac{
\displaystyle
\int_0^t
I_{\mathrm{AB}}(u)
\,du
}{
\displaystyle
\int_0^t
v_{\mathrm{AB}}(u)
\,du
}.
\label{eq:abi}
\end{equation}

A higher ABI indicates that perceptual orientation and physical action are assigned to different task areas for a larger proportion of the valid observation period. This descriptor captures coordination demand that is not visible from dwell time alone, such as visually monitoring the robot while continuing an assembly action.

\subsection{Context-aware Attention Switching Demand}
\label{subsec:asd_ctx}

The preceding descriptors characterize where the operator remains and whether perception and action are aligned. HRC assembly also requires operators to redirect attention between task contexts \cite{redick2016cognitive}. These transitions may impose different levels of reconfiguration demand and contribute to additional cognitive load \cite{MONSELL2003134}. The context-aware attention switching demand therefore evaluates confirmed transitions in $F(t)$, rather than raw frame-to-frame fluctuations.

A switching event is evaluated at each candidate state-update time and is counted only when the confirmed state changes between two task-related areas:

\begin{equation}
e_j
=
\mathbb{I}
\left[
F(t_j^-)
\neq
F(t_j^+)
\land
F(t_j^-)
\in
\mathcal{A}
\land
F(t_j^+)
\in
\mathcal{A}
\right],
\label{eq:switch_event}
\end{equation}

where $t_j$ denotes the $j$th candidate state-update time, and $t_j^-$ and $t_j^+$ represent the instants immediately before and after that update, respectively.

The contextual cost associated with a transition from area $a$ to area $b$ is defined as

\begin{equation}
c(a,b)
=
\begin{cases}
0,
& a=b,
\\[4pt]
c_{\mathrm{high}},
&
\{a,b\}
=
\{A_{\mathrm{ins}},A_{\mathrm{ass}}\},
\\[4pt]
c_{\mathrm{int}},
&
\{a,b\}
=
\{A_{\mathrm{col}},A_{\mathrm{id}}\},
\\[4pt]
c_{\mathrm{base}},
& \text{otherwise},
\end{cases}
\label{eq:transition_cost}
\end{equation}

subject to

\begin{equation}
c_{\mathrm{high}}
>
c_{\mathrm{int}}
>
c_{\mathrm{base}}
>
0.
\label{eq:cost_order}
\end{equation}

Instruction--assembly transitions receive the highest cost because they require reconfiguration between information interpretation and physical execution. Robot-collaboration--identification transitions receive an intermediate cost because robot-related events may interrupt object search or preparation. The remaining valid transitions are assigned the baseline cost. These values define a relative cost hierarchy rather than an absolute workload scale.

The context-aware attention switching demand is defined as the accumulated transition cost per unit of valid observation time:

\begin{equation}
\mathrm{ASD}_{ctx}(t)
=
\frac{C_{\mathrm{sw}}(t)}{T_v(t)}
=
\frac{
\displaystyle
\sum_{t_j\leq t}
e_j
c
\left(
F(t_j^-),
F(t_j^+)
\right)
}{
T_v(t)
}.
\label{eq:asd_context}
\end{equation}

Here, $C_{\mathrm{sw}}(t)$ denotes the accumulated context-weighted switching cost in the numerator. Each confirmed event contributes the transition-specific cost determined by its source and destination areas. Division by $T_v(t)$ expresses this accumulation as a context-weighted switching rate over valid observation time. Following the initialization convention defined above, $\mathrm{ASD}_{ctx}(t)$ is set to zero while $T_v(t)=0$.

Each newly confirmed transition produces an immediate increase whose magnitude depends on its contextual cost. Between switching events, continued growth of $T_v(t)$ gradually dilutes the influence of earlier transitions. The resulting indicator is reported as weighted switches per unit of valid observation time rather than as a bounded workload score.

By distinguishing transitions according to their source and destination areas, $\mathrm{ASD}_{ctx}$ extends a conventional switching-rate measure with task-context information. This formulation supports comparison of switching patterns across HRC periods with different valid observation durations.

\subsection{Operation Hesitation Demand}
\label{subsec:ohd}

Behavioral hesitation can serve as an observable manifestation of increased cognitive demand, because an operator may temporarily suspend movement while resolving uncertainty and interpreting task information \cite{bavelas1992interactive}. This cue is particularly relevant during human--robot interaction, where the operator must anticipate robot motion and coordinate the timing of shared actions \cite{HANEY2026104747}. Accordingly, the final descriptor, operation hesitation demand (OHD), characterizes sustained low-motion episodes during valid robot-related interaction periods, such as when the operator waits for assistance or prepares a subsequent collaborative action.

To identify sustained low-motion behavior, we first estimate the instantaneous hand velocity from consecutive filtered hand positions:
\begin{equation}
v_h(t)
=
\frac{
\left\|
\mathbf{x}_{m}(t)
-
\mathbf{x}_{m}(t-\Delta t)
\right\|
}{
\Delta t
},
\label{eq:hand_velocity}
\end{equation}
where $\mathbf{x}_{m}(t)$ denotes the filtered hand position and $\Delta t$ is the sampling interval. The resulting velocity is evaluated only within the relevant interaction context. Specifically, $C(t)\in\{0,1\}$ marks a valid robot-collaboration period, whereas $G(t)\in\{0,1\}$ indicates that the relevant operation remains incomplete. Combining these contextual gates with the low-motion condition gives the hesitation state
\begin{equation}
H(t)
=
\mathbb{I}
\left[
C(t)=1
\land
G(t)=1
\land
v_h(t)<v_{\mathrm{th}}
\right],
\label{eq:hesitation_state}
\end{equation}
where $v_{\mathrm{th}}$ is the low-motion threshold. Thus, $H(t)=1$ only when low hand velocity is observed during an active, incomplete collaboration. Observations outside collaboration or after operation completion are excluded before the hesitation duration is accumulated.

To quantify how long the gated low-motion state persists, its duration and the corresponding valid collaboration duration are accumulated as
\begin{align}
T_{\mathrm{hes}}(t)
&=
\int_0^t
H(u)
\,du,
\label{eq:hesitation_duration}
\\
T_{\mathrm{collab}}(t)
&=
\int_0^t
C(u)
\,du.
\label{eq:collaboration_duration}
\end{align}
Normalizing accumulated hesitation by valid collaboration time yields the operation hesitation demand
\begin{equation}
\mathrm{OHD}(t)
=
\frac{
T_{\mathrm{hes}}(t)
}{
T_{\mathrm{collab}}(t)+\varepsilon
},
\label{eq:ohd}
\end{equation}
where $\varepsilon>0$ is a small positive duration constant that prevents division by zero before the first valid collaboration period. A higher OHD value indicates that sustained low-motion episodes occupy a larger proportion of valid robot-interaction time. Because $H(t)$ already incorporates the collaboration and task-incompletion gates, ordinary inactivity outside these conditions does not contribute to the numerator.

\subsection{Composite HRC-CWL Output}
\label{subsec:composite_hrc_cwl}

After the task- and interaction-related descriptors have been obtained, their normalized values are integrated into the final HRC-CWL output. Let $\tilde{\mathbf{x}}(t)$ denote the normalized indicator vector comprising AED, TID, ICD, $\mathrm{ASD}_{\mathrm{ctx}}$, and OHD. ABI and robot-collaboration-area occupation are retained as diagnostic intermediate descriptors and are therefore not included in the final weighted aggregation. The composite assessment is defined compactly as

\begin{equation}
\mathrm{HRC\text{-}CWL}(t)
=
\mathbf{w}^{\mathrm{T}}\tilde{\mathbf{x}}(t),
\qquad
\mathbf{w}\succeq\mathbf{0},
\quad
\mathbf{1}^{\mathrm{T}}\mathbf{w}=1,
\label{eq:hrc_cwl}
\end{equation}

where $\mathbf{w}$ is the weight vector jointly determined from the relative importance ratings collected through the customized post-task questionnaire and expert judgments regarding task- and interaction-related demands in HRC assembly. The resulting weights are fixed across participants and experimental conditions. The composite output provides a concise workload-related summary, while the underlying indicators are retained to explain its task and interaction sources.


\section{Experimental Tests}
\label{sec:experimental_setup}

This section presents the experimental study conducted to validate the proposed assessment framework in a human--robot collaborative assembly scenario. The framework was evaluated under different task complexity levels to examine its ability to characterize cognitive workload variations and was further validated against behavioral, subjective, and physiological responses. The experimental procedure and validation methodology are described in the following subsections.

\subsection{Workspace Apparatus and Layout}


The experiment was conducted on a human--robot collaborative assembly platform consisting of a UR5e collaborative robot equipped with a parallel Robotiq gripper and an Intel RealSense D435i RGB-D camera. The robot delivered components to a predefined handover location, while the camera captured human--robot interactions from a calibrated workspace view. Task instructions were provided through a display interface, and behavioral and physiological data were synchronously recorded throughout the experiment.

\begin{figure}[h]
\centering
\includegraphics[width=\linewidth]{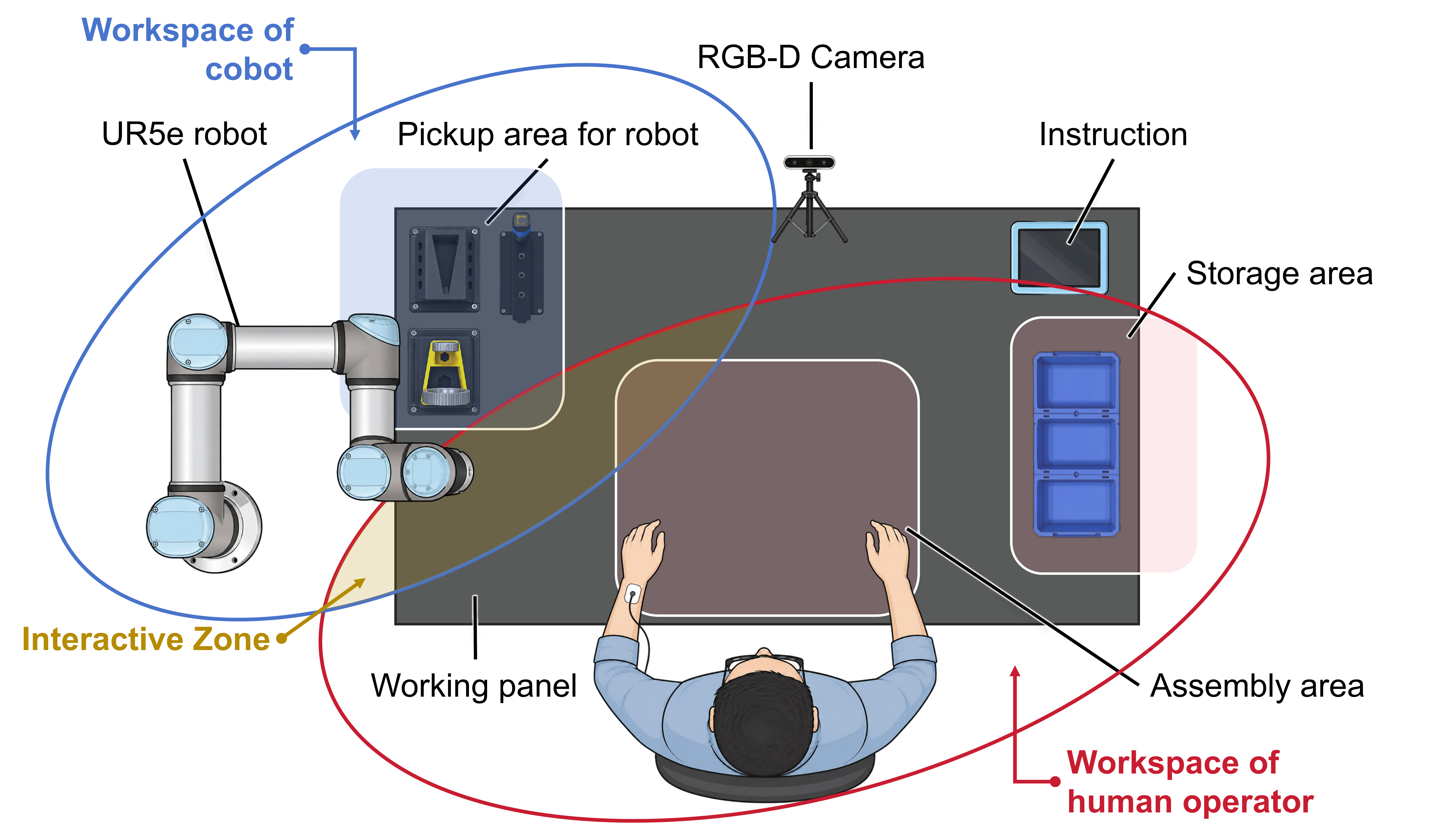}
\caption{Experimental setup for the human-robot collaborative assembly task.}
\label{fig:setup}
\end{figure}

As shown in Figure~\ref{fig:setup}, the workspace was organized into four task-related regions corresponding to the major information and interaction demands during assembly: storage, assembly, instruction, and robot collaboration areas. The storage and assembly regions supported component preparation and task execution, while the instruction region provided stepwise procedural guidance. The robot collaboration area covered the handover zone and the robot operating space where human--robot interaction occurred. All regions were registered within a unified workspace coordinate system, enabling consistent mapping of human attention, actions, and robot states during the experiment. 

The experimental scenario was designed to reflect a representative but simplified HRC workflow in industrial manufacturing. A gearbox assembly task was selected as a representative case to capture essential HRC characteristics while maintaining experimental controllability \cite{LUCCI2022102384}. By integrating robot-assisted component delivery with human assembly execution, the task enabled systematic investigation of how varying complexity levels influenced cognitive workload during collaborative assembly.


\subsection{Task Conditions}
Three collaborative assembly conditions with progressively increasing assembly, interaction, and collaboration complexity were designed based on the C-HRC complexity model proposed by Capponi et al.~\cite{CAPPONI2025103026} within a gearbox assembly scenario.

\begin{equation}
\begin{array}{rl}
C_{HRC} = &
\left[\dfrac{n_p}{N_p}+CI_{product}\right]\log_2(N_p+1)
+\left[\dfrac{n_s}{N_s}\right]\log_2(N_s+1)
\\[10pt]
&\quad
+\left[\dfrac{n_{rs}}{N_t}\right]\log_2(N_t+1)
+\dfrac{T_{H+R}}{T_C}\log_2(N_c+1).
\end{array}
\label{eq:chrc_complexity}
\end{equation}

\begin{figure}[h]
\centering
\includegraphics[width=0.6\linewidth]{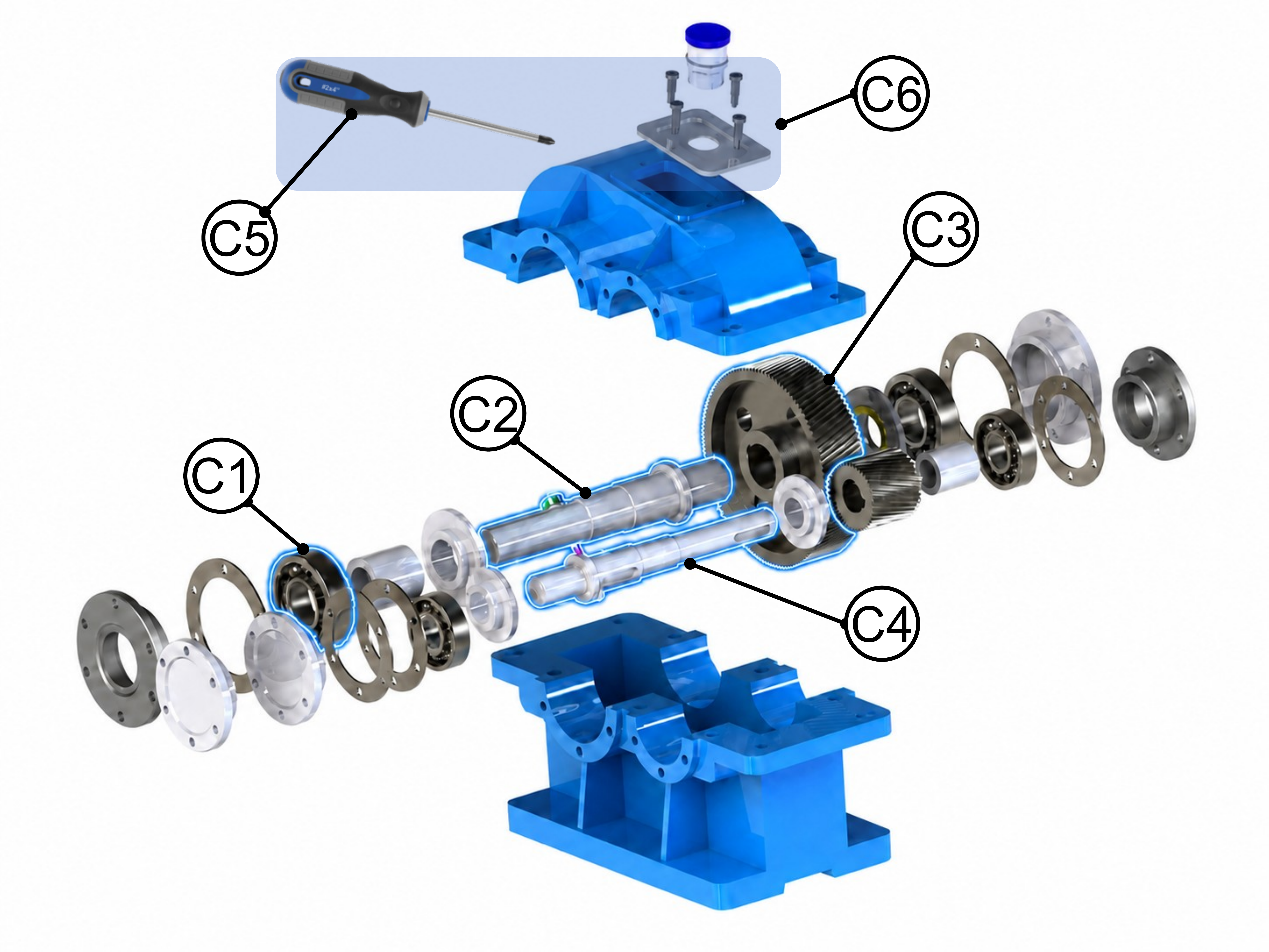}
\caption{Assembly steps and highlighted components to be delivered by robots}
\label{fig:sequence}
\end{figure}

In each condition, participants assembled a simplified gearbox from the base, cover, shafts, gears, bearings and fastening elements. The robot delivered selected components as highlighted in Figure \ref{fig:sequence}, requiring participants to integrate robot-introduced parts into the ongoing assembly. As task complexity increased, the conditions placed greater demands on assembly sequencing and robot monitoring, while preserving the same overall workflow (see Table~\ref{tab:task_design_complexity}).

During the experiment, the UR5e collaborative robot was configured through kinesthetic teaching to generate task-specific assistance trajectories for manipulating assembly components with diverse geometries, varying dimensions, and irregular shapes. The demonstrated trajectories were recorded and replayed during experiments, enabling consistent and repeatable robot assistance across different assembly operations.


\begin{table}[H]
\centering
\caption{Collaborative assembly tasks and score-derived difficulty profiles.}
\label{tab:task_design_complexity}
\small
\setlength{\tabcolsep}{6pt}
\renewcommand{\arraystretch}{1.15}
\resizebox{0.90\textwidth}{!}{%
\begin{tabularx}{\textwidth}{
    @{}
    >{\raggedright\arraybackslash}p{0.09\textwidth}
    >{\raggedright\arraybackslash}X
    >{\raggedright\arraybackslash}p{0.36\textwidth}
    @{}}
\toprule
\textbf{Level}
& \textbf{Task design}
& \textbf{Difficulty profile}
\tabularnewline
\midrule

\textbf{L1}
& The robot delivered one component from C1--C4, which the participant received and installed.
& Assembly complexity dominated, with relatively limited interaction and collaboration demands.
\tabularnewline
\addlinespace[4pt]

\textbf{L2}
& The robot delivered two components from C1--C4 and screwdriver C5 for assembly and fastening.
& The additional component and robot-supported fastening stage increased collaboration complexity relative to L1.
\tabularnewline
\addlinespace[4pt]

\textbf{L3}
& The robot delivered two components from C1--C4 and fastening module C6, comprising a screwdriver and screws, for assembly and fastening.
& The integrated module required coordination with a broader set of robot-provided resources, further increasing collaboration complexity relative to L2.
\tabularnewline

\bottomrule
\end{tabularx}
}
\end{table}

\subsection{Participants}
All procedures involving human participants were reviewed and approved by the institutional ethics review board of the university under approval HKUST(GZ)-HSP-2026-0088. A total of ten participants (average age: $21.6 \pm 2.0$ years) were recruited from the university community on a voluntary basis. The inclusion criteria required participants to have normal or corrected-to-normal vision and no extensive prior experience with the specific assembly task, ensuring that workload responses were evaluated under comparable task familiarity. Before the experiment, participants were informed of the study procedure and provided written informed consent. Each participant completed all three task levels in a within-subject design, enabling comparisons of workload responses under different levels of task complexity.


\subsection{Experimental Procedure}

\begin{figure}[h]
\centering
\includegraphics[width=\linewidth]{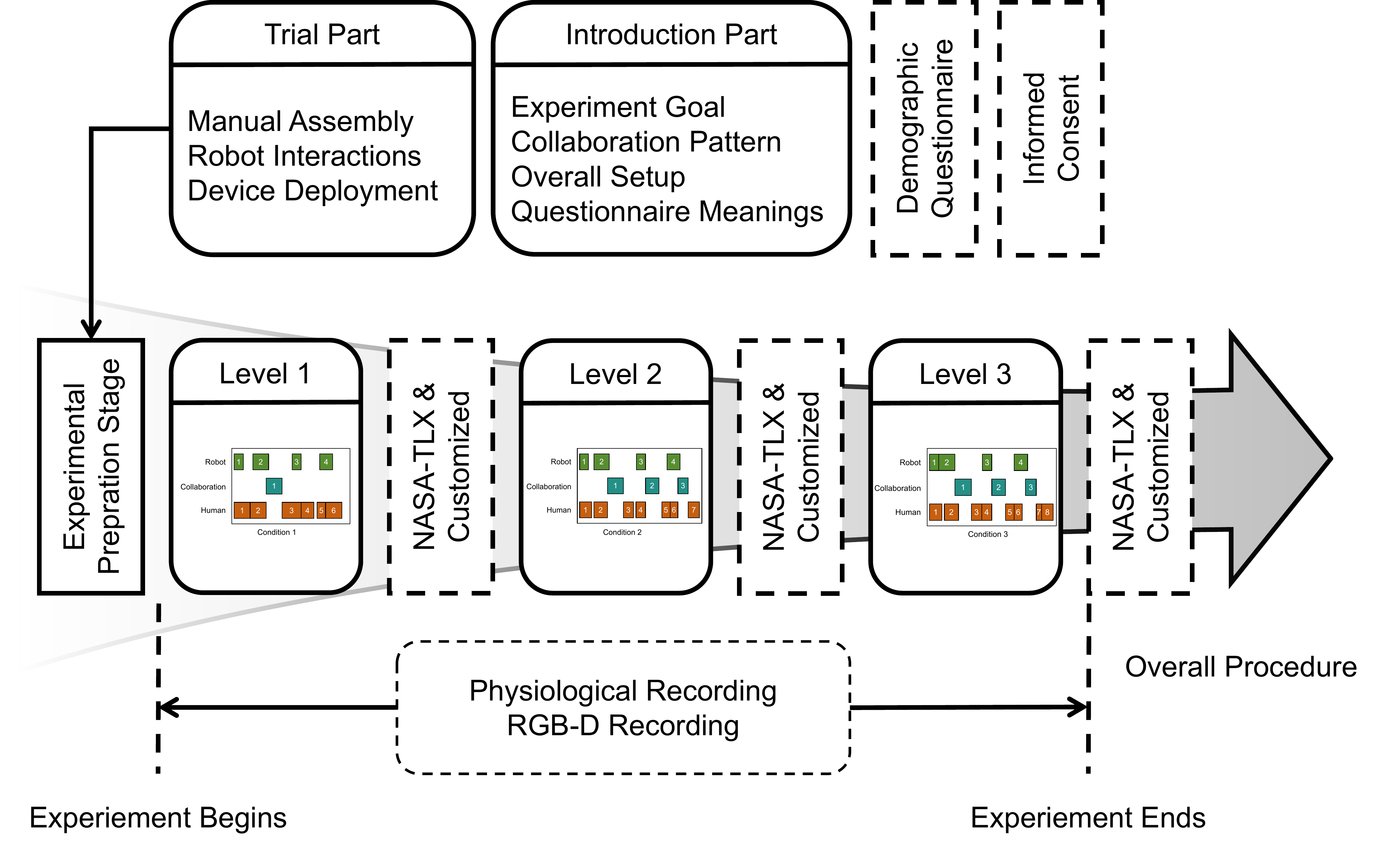}
\caption{Overall experimental procedure.}
\label{fig:procedure}
\end{figure}

The overall experimental procedure is shown in Figure~\ref{fig:procedure}. Before the formal trials, participants were briefed on the study aims and familiarised with the collaborative assembly environment, including the workstation layout, robot handover process, assembly procedure and sensing setup. Demographic information and written informed consent were obtained before any experimental recording.

Participants then completed a practice session to familiarize themselves with the assembly task, robot handover process and instruction interface. After a short baseline recording, they performed the three HRC assembly tasks in a randomized order, with task complexity defined by Levels 1, 2 and 3. Subjective workload was assessed after each condition using the NASA-TLX questionnaire, whereas the customized post-task questionnaire was completed after all three experiments (the custom survey is provided in \ref{app1}). Physiological signals and visual streams were recorded continuously throughout the experiment.

\subsection{Data Collection}

\begin{figure}[H]
    \centering
    \includegraphics[width=\linewidth]{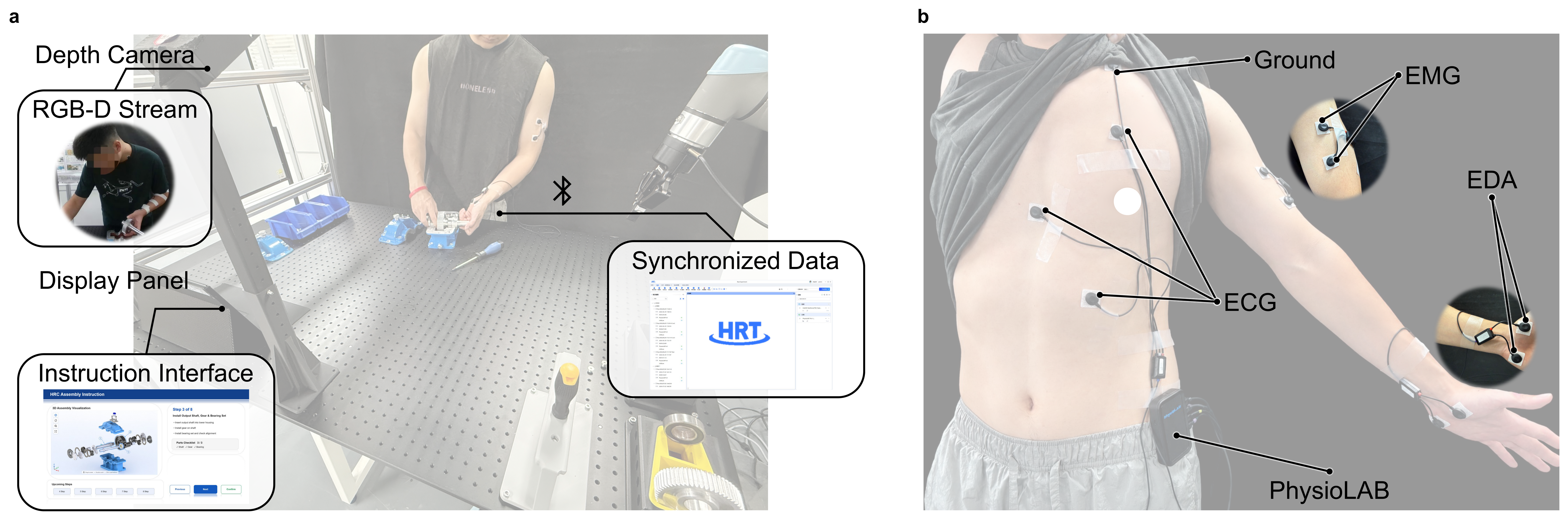}
    \caption{
    Experimental instrumentation and data collection setup.
    \textbf{a}, Human--robot collaborative assembly environment and synchronized data acquisition system.
    \textbf{b}, Physiological sensing configuration, including ECG, EMG and EDA.
    }
    \label{fig:sensor-setup-environment}
\end{figure}

The data collection setup is shown in Figure~\ref{fig:sensor-setup-environment}. During the experiment, the recording system continuously captured visual (both attentional and behavioral), physiological, and robot-related data to characterize operator states and human--robot interactions. RGB video acquired by an Intel RealSense D435i camera was used to extract visual attention and upper-body behavioral cues, which supported the proposed vision-based cognitive workload assessment framework. Robot-side measurements provided information on end-effector motion and interaction states during collaborative assembly. Physiological responses, including ECG, EMG, and EDA signals, were recorded using PhysioLAB Pro1 sensors at 1000 Hz and served as reference measures for validating workload variations under different task complexity levels. All data streams were temporally synchronized using the Human Research Tool (HRT; Info Instrument) and stored according to confidentiality requirements.

\section{Experimental Results}
\label{sec:experimental_results}

The evaluation first examines the temporal behavior of the proposed indicators, followed by comparisons across task conditions and correlation analysis with independent physiological measures. It then demonstrates the real-time deployment of the framework in the collaborative assembly setting.

\subsection{Statistical Comparison Across Task Conditions}
\label{subsec:statistical_comparison}

The temporal observations were subsequently evaluated at the group level to determine whether workload-related measures differed systematically across the three task complexity conditions. The comparison provides the statistical basis for distinguishing condition-dependent changes from variability among participants and repeated trials.

\begin{figure}[H]
    \centering
    \includegraphics[width=\textwidth]{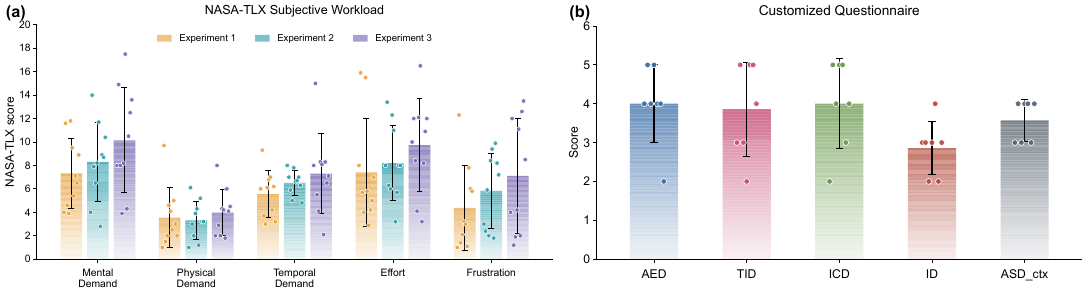}
    \caption{
    Quantitative visualization of subjective questionnaire results.
    \textbf{(a)} NASA-TLX subjective workload profiles across experimental conditions, where solid lines indicate mean scores and shaded regions represent variability.
    \textbf{(b)} Customized questionnaire scores across five metrics, where bars indicate mean scores and error bars indicate standard deviation.
    }
    \label{fig:questionnaire-combined}
\end{figure}

Subjective workload was quantified using the unweighted raw NASA-TLX ratings collected from all ten participants (Figure~\ref{fig:questionnaire-combined}a). The overall raw score was calculated as the arithmetic mean of the six dimension ratings and increased from 6.75 in Condition~1 to 7.26 in Condition~2 and 8.45 in Condition~3. A Friedman test detected a significant condition effect on this overall score ($\chi^2(2) = 6.20$, $p = 0.045$). At the dimension level, mental demand increased from 7.33 to 8.31 and 10.14, while temporal demand rose from 5.55 to 6.51 and 7.30. Significant condition effects were observed for both mental demand ($\chi^2(2) = 6.70$, $p = 0.035$) and temporal demand ($\chi^2(2) = 6.82$, $p = 0.033$). Effort and frustration also increased descriptively, from 7.40 to 9.74 and from 4.37 to 7.10, respectively, but neither effect was significant (effort: $\chi^2(2) = 5.42$, $p = 0.067$; frustration: $\chi^2(2) = 4.20$, $p = 0.122$). Physical demand ($\chi^2(2) = 2.80$, $p = 0.247$) and performance ($\chi^2(2) = 1.00$, $p = 0.607$) also showed no significant condition effects. These findings indicate that increasing task complexity was associated with higher overall perceived workload, with the clearest changes occurring in mental and temporal demand.

\subsection{Temporal Dynamics of Model Indicators}
\label{subsec:indicator_dynamics}

\begin{figure}[H]
    \centering
    \includegraphics[width=0.9\textwidth]{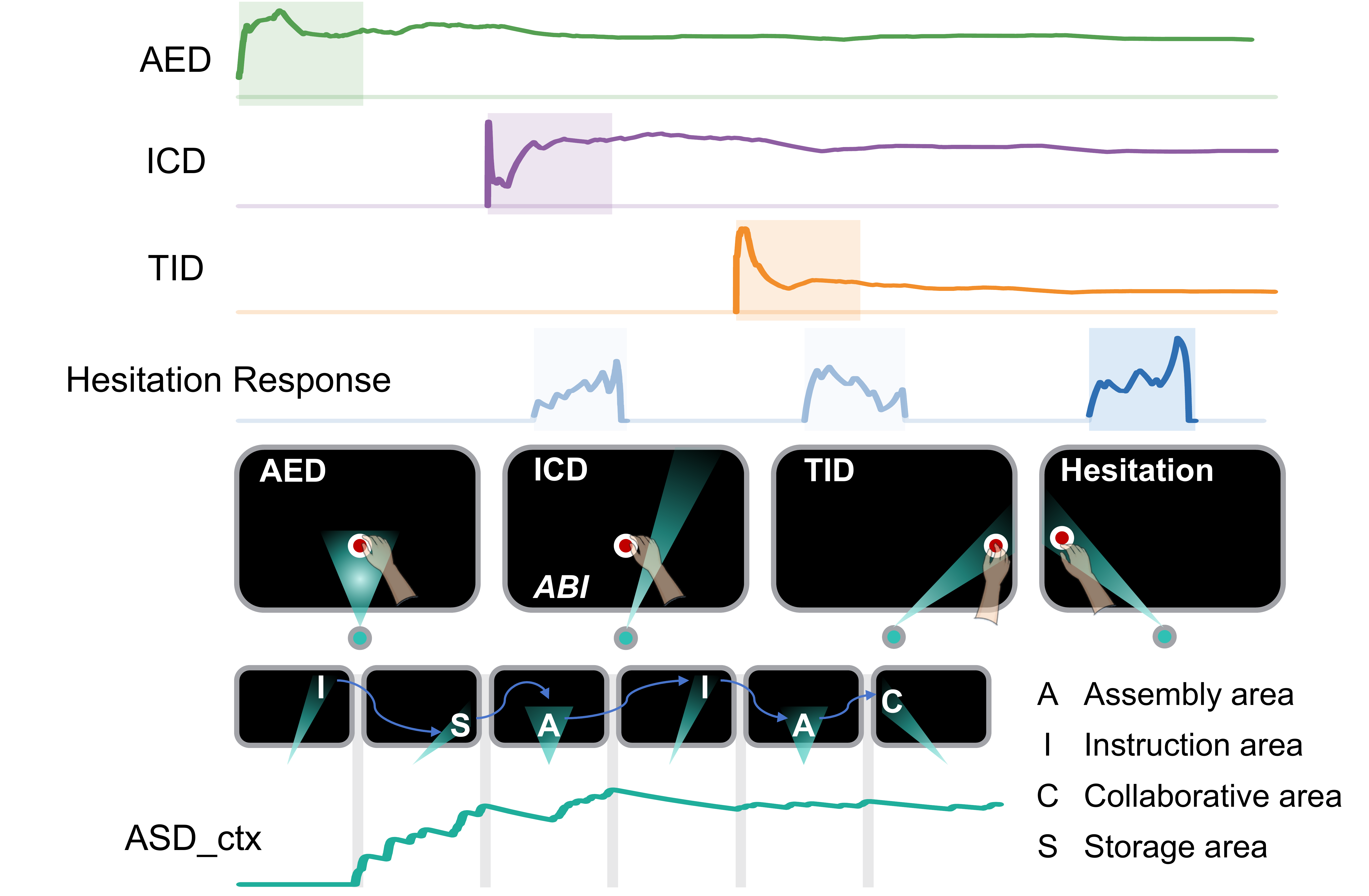}
    \caption{
    Visualization of the temporal indicators derived from confirmed area allocation, context transitions, and hesitation-related behavior.
    }
    \label{fig:model-metrics-explanation}
\end{figure}

The temporal outputs were derived from confirmed attention--action states. Assigning fused attention to a calibrated task area activated the corresponding area-specific indicator, whose trajectory showed the accumulated allocation to that context. Confirmed transitions between areas generated stepwise changes in $ASD_{\mathrm{ctx}}$, whereas sustained low hand motion during unresolved collaboration generated localized OHD responses. Figure~\ref{fig:model-metrics-explanation} integrates this mapping from spatial attention and hand behavior to the resulting temporal trajectories.

\subsubsection{Area-specific Attentional Dynamics}
\label{subsec:attentional_level_indicators}

Figure~\ref{fig:p08-attentional-demand} shows the temporal evolution of the area-specific attentional demands. After brief initial fluctuations, AED remained above ICD and TID for most of the task and increased modestly, whereas the latter indicators progressively declined. Their presentation on a common timeline reveals how the relative allocation of task-directed attention developed as assembly proceeded.

\subsubsection{Switching and Hesitation Dynamics}
\label{subsec:action_level_indicators}

\begin{figure}[H]
    \centering
    \begin{subfigure}[t]{0.48\textwidth}
        \centering
        \includegraphics[width=\textwidth]{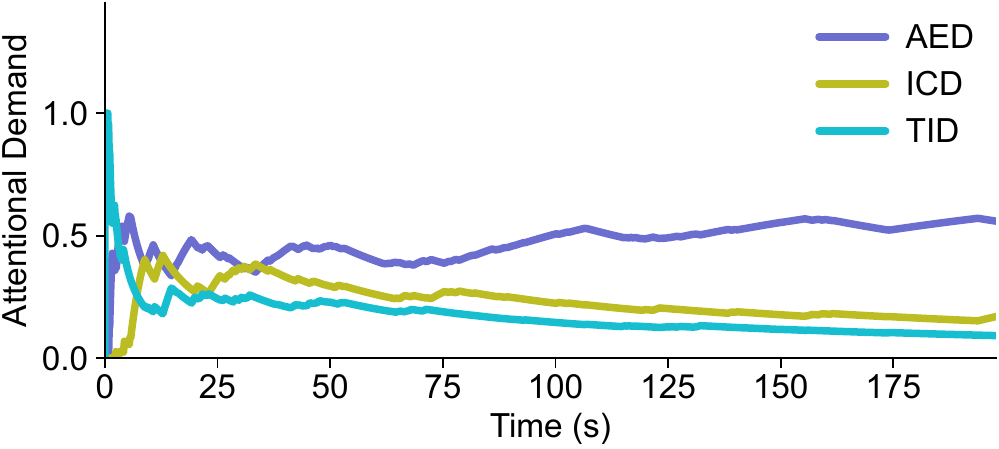}
        \caption{\textbf{(a)} Area-specific attentional demands.}
        \label{fig:p08-attentional-demand}
    \end{subfigure}
    \hfill
    \begin{subfigure}[t]{0.48\textwidth}
        \centering
        \includegraphics[width=\textwidth]{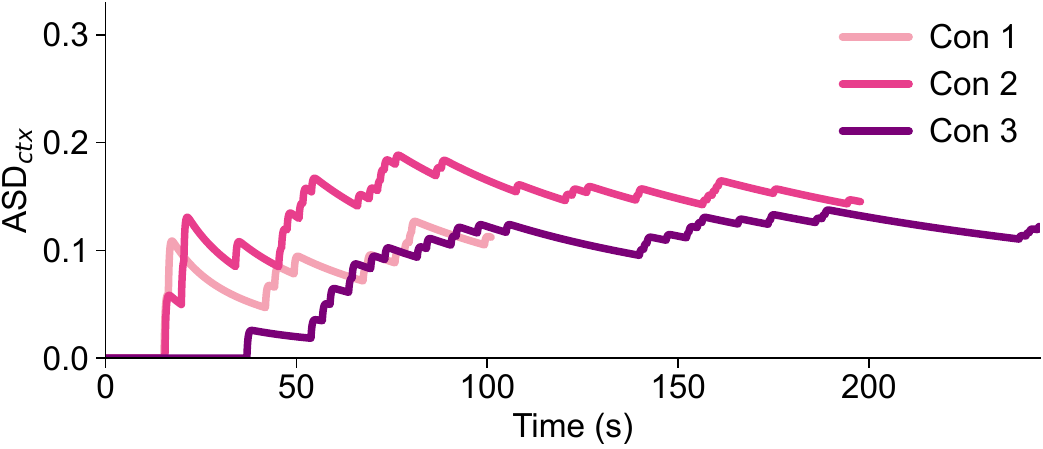}
        \caption{\textbf{(b)} Context-aware switching demand.}
        \label{fig:p08-asdctx-trend}
    \end{subfigure}
    \par\medskip
    \begin{subfigure}[t]{0.48\textwidth}
        \centering
        \includegraphics[width=\textwidth]{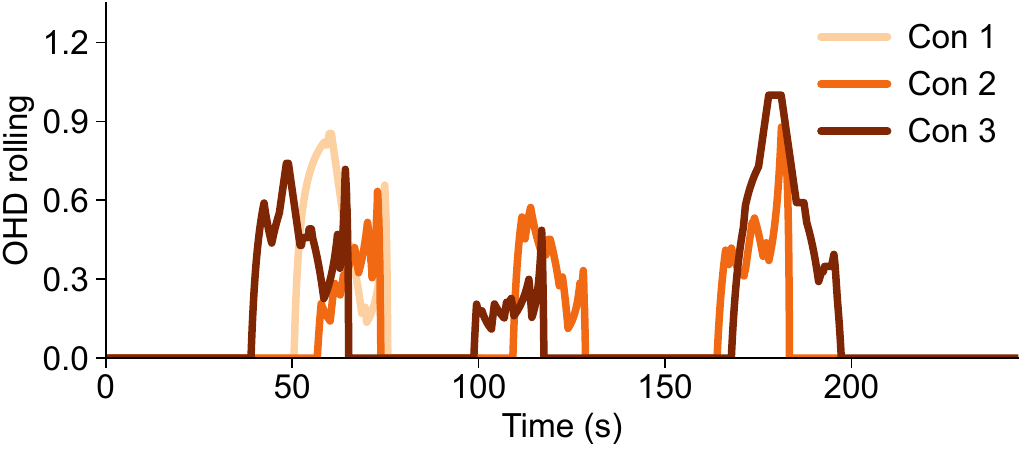}
        \caption{\textbf{(c)} Hesitation.}
        \label{fig:p08-ohd-trend}
    \end{subfigure}
    \caption{Representative temporal dynamics of the model indicators during human--robot collaborative assembly. \textbf{(a)} AED, ICD, and TID show how assembly execution, instruction checking, and tool identification demands evolve during task execution. \textbf{(b)} Stepwise increases in $ASD_{\mathrm{ctx}}$ correspond to confirmed context transitions, whereas subsequent decreases reflect temporal dilution in the absence of additional switching events. \textbf{(c)} Nonzero OHD segments indicate sustained low-motion behavior detected within valid robot-collaboration periods. These trajectories illustrate within-trial indicator behavior and are not intended as group-level comparisons.}
    \label{fig:p08-indicator-dynamics}
\end{figure}

Panels~b and~c of Figure~\ref{fig:p08-indicator-dynamics} show the event-dependent dynamics of attention switching and operation hesitation. Confirmed context transitions produced stepwise increases in $ASD_{\mathrm{ctx}}$, followed by gradual decreases when no additional switching event occurred. OHD instead appeared as discrete nonzero intervals when sustained low hand motion was detected during an active, incomplete collaboration. The three conditions are overlaid to compare the timing and persistence of these temporal patterns rather than to support group-level inference.

\subsection{Physiological Correlation Analysis}
\label{subsec:physiological_correlation}

Physiological signals were used as independent references to examine whether the vision-derived outputs corresponded with workload-related responses at event and segment scales. Event-level synchronization assessed localized physiological changes during hesitation episodes, whereas segment-level analysis examined associations across task progression and experimental conditions.

For the event-level analysis, the EDA signal was decomposed into tonic and phasic components:
\begin{equation}
EDA(t)=SCL(t)+SCR(t)+\epsilon(t),
\end{equation}
where $SCL(t)$ represents tonic arousal, $SCR(t)$ captures transient sympathetic responses, and $\epsilon(t)$ denotes residual activity. Mean SCL, SCR peak count, mean SCR amplitude, and phasic activity were extracted to characterize arousal over the synchronized windows.

\begin{figure}[htbp]
    \centering
    \includegraphics[width=0.9\textwidth]{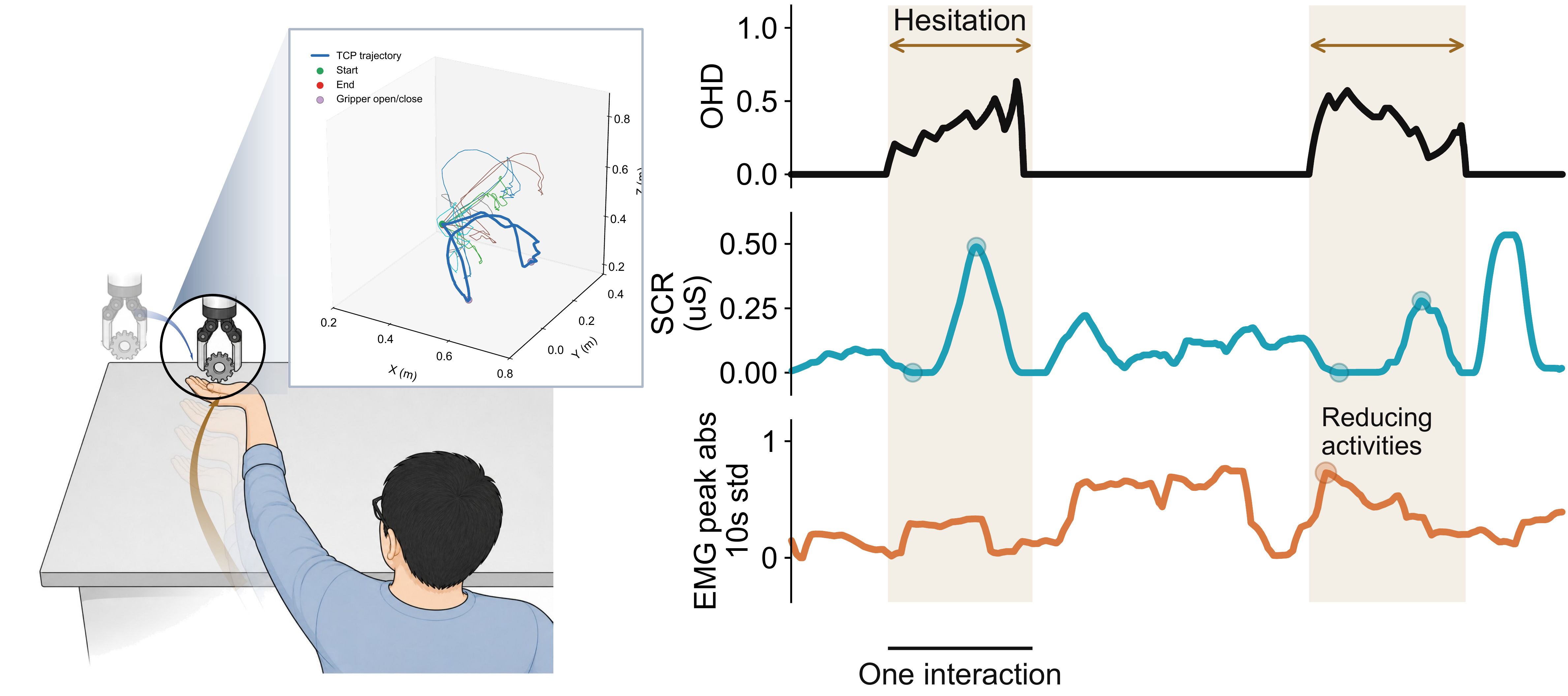}
    \caption{Representative robot-related interaction combining the spatial interaction context with synchronized OHD, SCR, and EMG trajectories. Shaded intervals denote hesitation periods detected by the assessment framework.}
    \label{fig:interaction-ohd}
\end{figure}

Figure~\ref{fig:interaction-ohd} presents a representative robot-related interaction in which OHD is aligned with SCR and EMG activity. Two shaded hesitation intervals contained nonzero OHD responses together with localized SCR and EMG changes. Additional physiological variation outside these intervals indicates that the correspondence was temporal but not one-to-one. The comparison therefore illustrates event-level alignment without implying causal coupling.

\begin{figure}[H]
\centering
\includegraphics[width=\linewidth]{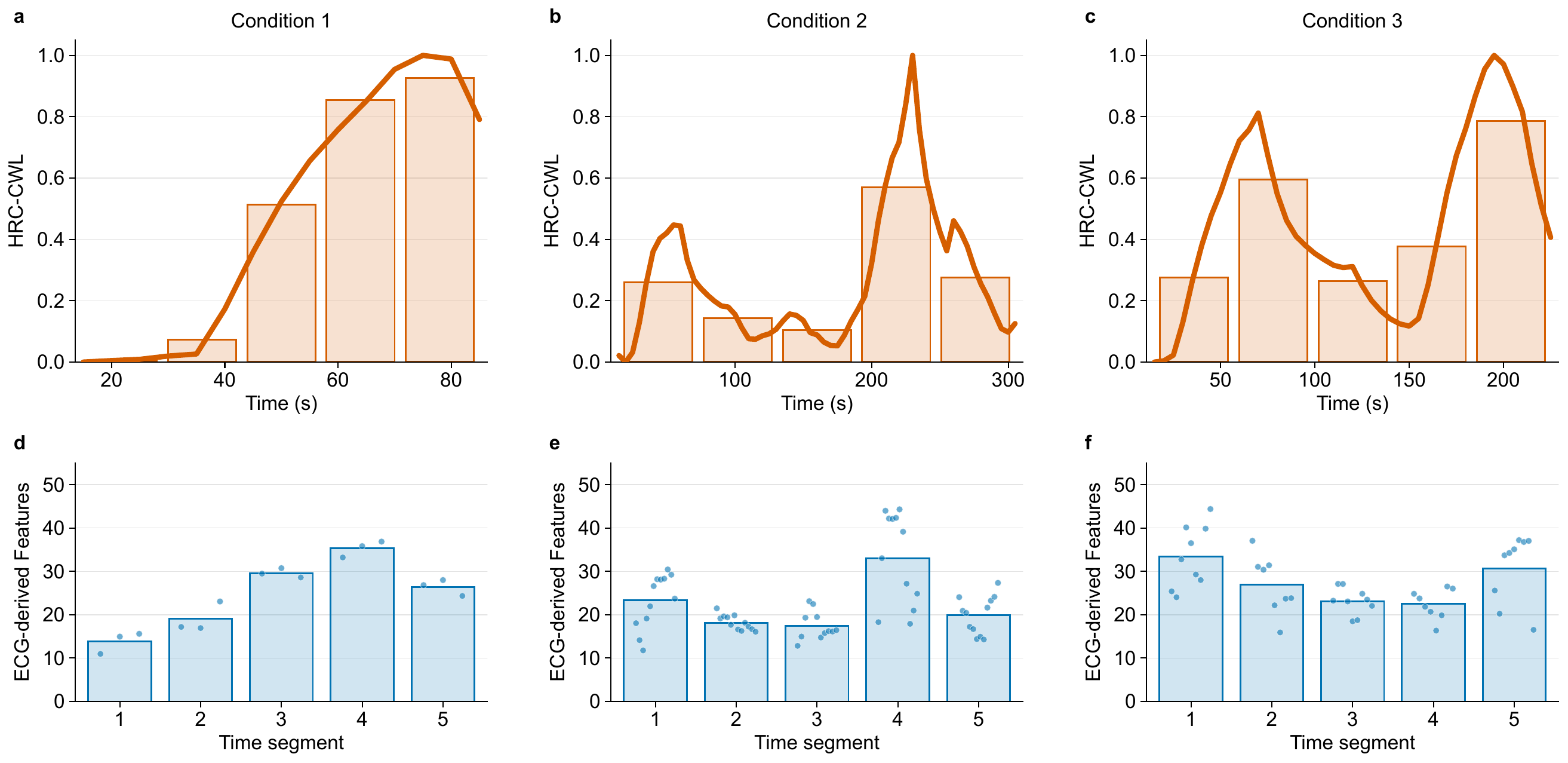}
\caption{Segment-level comparison between HRC-CWL and ECG-derived features across the three collaborative assembly conditions.} 
\label{fig:task_level_trend}
\end{figure}

At the broader temporal scale, Figure~\ref{fig:task_level_trend} compares HRC-CWL with ECG-derived features across five synchronized task segments in each condition. Both outputs varied across the segments, but their degree of correspondence differed between conditions and was not uniformly monotonic. This heterogeneity motivated participant-specific correlation tests instead of a pooled trend comparison.

Spearman's rank correlation tested the monotonic association between HRC-CWL and ECG-derived features across synchronized observations for each participant. Table~\ref{tab:spearman_hrccwl_sdnn} reports only participant-specific $p$ values, allowing significance to be assessed without implying an unreported direction or effect size.

\begin{table}[H]
\centering
\caption{Participant-level Spearman rank-correlation tests between HRC-CWL and ECG-derived features across synchronized observations.}
\label{tab:spearman_hrccwl_sdnn}
\small
\renewcommand{\arraystretch}{1.18}
\begin{tabular*}{0.86\textwidth}{@{\extracolsep{\fill}}cccc@{}}
\toprule
\textbf{Participant} & \textbf{$p$ value} & \textbf{Participant} & \textbf{$p$ value} \\
\midrule
P01 & 0.00014\textsuperscript{***} & P06 & -0.5986 \\
P02 & 0.5115 & P07 & $1.97\times10^{-5}$\textsuperscript{***} \\
P03 & 0.00045\textsuperscript{***} & P08 & $4.41\times10^{-6}$\textsuperscript{***} \\
P04 & $3.43\times10^{-5}$\textsuperscript{***} & P09 & 0.0027\textsuperscript{**} \\
P05 & 0.0063\textsuperscript{*} & -- & -- \\
\bottomrule
\end{tabular*}
\vspace{3pt}

\begin{minipage}{0.86\textwidth}
\footnotesize
\textit{Note:} Entries are participant-specific $p$ values for the Spearman correlation between HRC-CWL and ECG-derived features across synchronized observations. Asterisks denote \textsuperscript{*}$p<0.05$, \textsuperscript{**}$p<0.005$, and \textsuperscript{***}$p<0.001$; unmarked values are not significant. Correlation direction and magnitude are not reported.
\end{minipage}
\end{table}

For seven of the nine participants, Spearman's test identified a statistically significant association between HRC-CWL and ECG-derived features (Table~\ref{tab:spearman_hrccwl_sdnn}). The remaining two participants showed no significant association, indicating that the correspondence was present in most participants but was not uniform across the sample.


Taken together, these analyses showed that the physiological correspondence of the vision-derived outputs varied with temporal scale and among participants.

\subsection{Real-Time Deployment in HRC Assembly}
\label{subsec:online_implementation}

The assessment framework was deployed as a real-time monitoring pipeline in the collaborative assembly scenario. Incoming RGB-D observations were combined with synchronized robot states and interpreted within the calibrated task-related areas. Visual evidence from operator orientation and hand position was fused and temporally confirmed before the attention--action state was updated. This procedure reduced frame-level fluctuations while preserving sustained behavioral changes relevant to the ongoing task. Importantly, when deployed in practical HRC settings, the framework provides real-time vision-based workload assessment without requiring operators to wear any additional sensors.

\begin{figure}[H]
    \centering
    \includegraphics[width=\textwidth]{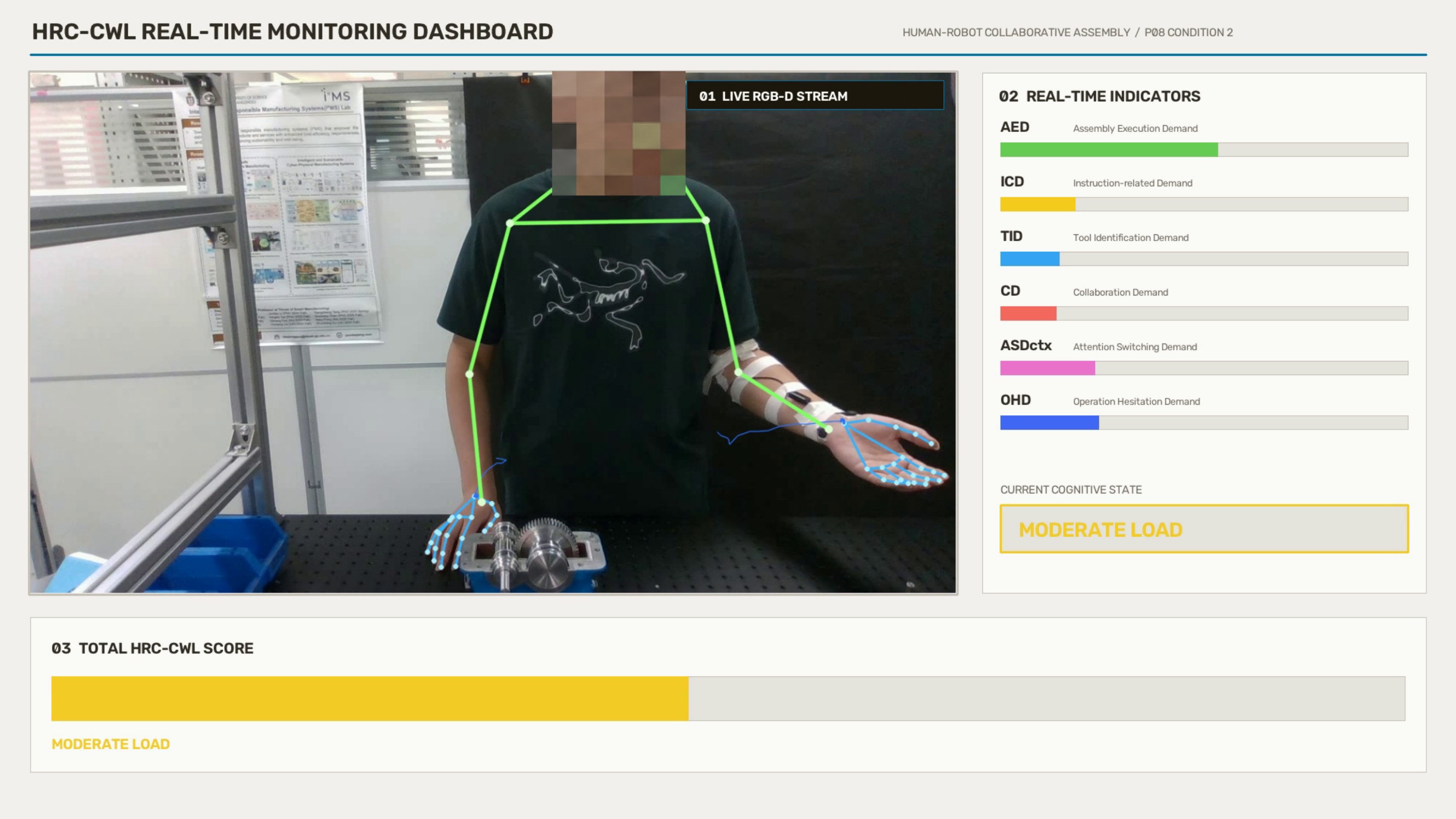}
    \caption{Real-time HRC-CWL monitoring interface in collaborative assembly. The representative view shows one participant performing the HRC task. The left panel presents the live RGB-D stream with detected facial, upper-body, and hand landmarks. The right panel presents the workload-related indicators generated by the framework in real time. The lower panel displays the interface-level composite HRC-CWL score and its corresponding workload category. Displaying these indicators separately helps identify which task or interaction behavior is associated with a change in the overall output.}
    \label{fig:hrc-cwl-implementation}
\end{figure}

The deployed interface is shown in Figure~\ref{fig:hrc-cwl-implementation}. It displayed the live RGB-D stream with detected landmarks alongside the area-related demand indicators, $\mathrm{ASD}_{\mathrm{ctx}}$, and OHD. The individual indicators remained visible beside the composite $\mathrm{HRC\text{-}CWL}(t)$ output and its workload category, allowing the summary to be traced to its task and interaction sources. The displayed category was used as an interface-level summary rather than a direct psychophysiological measurement. Representative demonstrations of this real implementation are provided in the \ref{app1}.

\section{Discussion}

The findings suggest that workload-related demand in collaborative assembly is expressed through how operators distribute attention across task contexts and coordinate their actions with the robot. By mapping temporally confirmed attention--action states onto functionally defined workspace areas, the framework reveals how demand emerges during task execution and human--robot interaction. This process-oriented representation is particularly relevant when visual monitoring of the robot occurs alongside ongoing manual activity, a coordination pattern that cannot be fully explained by an overall workload rating alone. The proposed indicators therefore provide an interpretable account of where demand is concentrated and how it develops during collaboration.

The physiological comparisons provide convergent rather than definitive support for the vision-derived assessment. Associations between HRC-CWL and ECG-derived features indicate correspondence at the task-segment level, while the alignment of OHD with SCR and EMG responses suggests sensitivity to localized interaction episodes. Because these relationships varied among participants and were not one-to-one, the outputs should be interpreted as behavioral proxies rather than direct psychophysiological measurements of cognitive workload.

The practical value of the framework lies in combining interpretable assessment with lightweight real-time deployment. Extending vision-based workload assessment toward HRC-specific attention and coordination demands \cite{9795908}, the method generates continuous workload-related metrics from a single RGB-D stream and synchronized robot states without requiring participant-mounted sensors. Rather than relying solely on a single workload score, it retains distinct indicators that can help determine whether an increase is associated with assembly execution, information seeking, attention switching, or hesitation. This diagnostic structure provides a basis for subsequent system-level optimization, such as adjusting robot motion, handover timing, or task guidance according to the inferred source of demand.

The present findings were obtained from a small participant sample in a single calibrated assembly setting, which limits their generalizability. The use of fixed indicator weights may also restrict the framework's ability to accommodate individual and contextual differences. Future research should therefore validate the assessment in other collaborative environments before examining whether indicator-guided robot adaptation can improve collaboration.

\section{Conclusion}

This study developed a vision-based attention--action assessment framework for characterizing workload-related demands in human--robot collaborative assembly. Rather than relying solely on a single workload score, the framework fuses head, upper-body, and hand cues with robot-interaction context within calibrated task-related areas to construct a temporally confirmed attention--action state. The resulting indicators distinguish where task demand is concentrated and how it develops through area-specific task demand, attention--behavior inconsistency, context-aware attention switching, and operation hesitation.

The framework was evaluated in a three-level collaborative gearbox assembly experiment through temporal indicator analysis, group-level task-condition comparison, synchronized physiological analysis, and online implementation. The temporal results showed that the indicators captured distinct patterns of task-context reallocation and robot-related hesitation. Comparisons with ECG- and EDA-derived features provided preliminary evidence that the model outputs were consistent with workload-related physiological variation, while the synchronized analysis of OHD, SCR, and EMG showed that this correspondence was localized to specific interaction episodes. The online implementation further demonstrated that the composite HRC-CWL output and its underlying indicators could be generated and traced during task execution.

Future work will extend the proposed real-time vision-based assessment framework to support human digital twin (HDT) applications in collaborative manufacturing \cite{Wang2022HDT}. By continuously mapping workload-related changes in operator behavior, the framework can support cognitive representations of human operators in digital twin environments. These representations may further enable human-centered monitoring and adaptive robot assistance by guiding collaborative behavior adjustment according to the operator's real-time state.

\section*{CRediT authorship contribution statement}

\textbf{Junyan Xiong:} Conceptualization, Data curation, Investigation, Methodology, Software, Validation, Formal analysis, Writing -- original draft. \textbf{Naiyi Feng:} Conceptualization, Investigation, Writing -- review \& editing. \textbf{Xingke Xia:} Methodology, Validation, Visualization. \textbf{Qihang Fan:} Visualization, Writing -- review \& editing. \textbf{Suchang Chen:} Validation. \textbf{Daqiang Guo:} Conceptualization, Methodology, Writing -- review \& editing, Supervision, Funding acquisition, Project administration.




\section*{Declaration of Generative AI and AI-assisted technologies in the writing process}

During the preparation of this work, the authors used OpenAI's GPT-5.6 to enhance the readability and language of the manuscript. To improve the clarity and overall visual presentation of the schematic figures, OpenAI's GPT-5.6 was also used to generate selected illustrative elements; however, all content relating to the model indicators was based on data collected in the experiments. After using these tools, the authors reviewed and edited the content as needed and take full responsibility for the content of the published article.

\section*{Declaration of competing interest}

The authors declare that they have no known competing financial interests or personal relationships that could have appeared to influence the work reported in this paper.

\section*{Data availability}

Data will be made available on request.


\appendix
\section{Supplementary Materials}
\label{app1}
Additional demonstration videos, experimental recordings, and sample results are provided in the supplementary ZIP file. These supplementary materials include representative demonstrations of the human–robot collaborative assembly experiments, original experimental recordings, and data visualizations used in the framework analysis. They are provided to enhance transparency and facilitate reproducibility, independent evaluation, and further investigation of the proposed HRC cognitive workload assessment framework.





\bibliographystyle{elsarticle-num}
\bibliography{references}




\end{document}